\documentclass{llncs}
\usepackage[utf8]{inputenc}
\usepackage[T1]{fontenc}
\usepackage{helvet}
\usepackage{pdfpages}
\usepackage{graphicx}% http://ctan.org/pkg/graphicx
\usepackage[footnotesize,hang]{caption} % réduire la taille des légendes des images
\usepackage{hyperref} % liens cliquables
\usepackage{import}
\usepackage{svg} % gérer les .svg directement
\usepackage{tikz} % du tikz c'est bien
\usetikzlibrary{positioning}
\usetikzlibrary{patterns}
\usetikzlibrary{calc}
\usetikzlibrary{shapes,arrows}
\usepackage{xcolor} % plus de couleurs
\usepackage{listings} % inline verbatim
\usepackage{amsmath} % des maths
\usepackage{amssymb} % plus de maths
\usepackage{pifont} % symboles en plus, genre x et v
\usepackage{cleveref} %références avec bon acronyme
\crefname{algocf}{alg.}{algs.}
\newcommand\restr[2]{{% we make the whole thing an ordinary symbol
  \left.\kern-\nulldelimiterspace % automatically resize the bar with \right
  #1 % the function
  \vphantom{\big|} % pretend it's a little taller at normal size
  \right|_{#2} % this is the delimiter
  }}
\usepackage{array}
\usepackage{xspace}
\newcolumntype{L}[1]{>{\raggedright\let\newline\\\arraybackslash\hspace{0pt}}m{#1}}
\newcolumntype{C}[1]{>{\centering\let\newline\\\arraybackslash\hspace{0pt}}m{#1}}
\newcolumntype{R}[1]{>{\raggedleft\let\newline\\\arraybackslash\hspace{0pt}}m{#1}}
\date{}
\usepackage[linesnumbered,lined,boxed,noend]{algorithm2e} % algorithmes et pseudocode

\SetCommentSty{mycommfont} % commentaires en bleu
\usepackage{inconsolata} %font for functions outside verb
\usepackage[sorting=none]{biblatex}
\newcommand{\norm}[2]{\left\lVert#1\right\rVert_{#2}}

\newcommand{\eg}{\emph{e.g.,}\xspace}

\newcommand{\state}{\mathcal{S}\xspace}%state of set of neurons
\newcommand{\facet}{\mathcal{F}\xspace}%facet
\newcommand{\relu}{ReLU\xspace}

\title{DISCO Verification: Division of Input Space
into COnvex polytopes for neural network verification}

\authorrunning{Girard-Satabin, Varasse, Schoenauer, Charpiat and Chihani}
\titlerunning{DISCO}
\author{Julien Girard-Satabin\inst{1}\inst{2}
    \and Aymeric Varasse\inst{1}
    \and Marc Schoenauer\inst{2} 
    \and Guillaume Charpiat\inst{2}
    \and Zakaria Chihani\inst{1}
}
  \institute{
      Université Paris-Saclay, CEA, List, F-91120, Palaiseau, France\\
      \email{julien.girard2@cea.fr, aymeric.varasse@cea.fr,
      zakaria.chihani@cea.fr}\medskip\\
      \and
      TAU team, LISN (Université Paris-Saclay and CNRS), INRIA\\
  \email{guillaume.charpiat@inria.fr, marc.schoenauer@inria.fr}}
\begin{document}

\maketitle

% \vspace{3cm}
% % Abstract
\begin{abstract}
The impressive results of modern neural networks partly come from their
non linear behaviour.
Unfortunately, this property makes it very difficult to apply
formal verification tools, even if we restrict ourselves to networks
with a piecewise linear structure.
However, such networks yields subregions that are linear and thus
simpler to analyse
independently. In this paper, we propose a method
to simplify the verification problem by operating a
partitionning into multiple linear subproblems. To evaluate the feasibility
of such an approach, we perform an empirical analysis of neural networks
to estimate the number of linear regions, and compare them to the bounds
currently known. We also present the impact of a technique aiming
at reducing the number of linear regions during training.
\end{abstract}

% Introduction and related work
\section{Introduction}

Over the last years, the class of programs known as deep neural networks has
been the topic of considerable work. Known to be theoretically able
to approximate any function with sufficiently many neurons,
their ability to process highly
dimensional inputs (speech, images, videos\dots) only guided with labeled
examples paved the way to multiple real-world applications.
However, as programs, deep neural networks are not exempt of malfunctions,
and research exhibited quite a few. Adversarial examples are
human-imperceptible, voluntary perturbations of the input that result in a
wrong answer of the program. They
can be found on multiple kinds of perceptual inputs
(images, audio~\cite{qinImperceptibleRobustTargeted2019},
video~\cite{chenShapeShifterRobustPhysical2018}),
and even be transferred between programs~\cite{papernotTransferabilityMachineLearning2016};
currently known countermeasures do not soundly prevent adversarial
examples~\cite{tsiprasRobustnessMayBe2018}.
It was also shown that it is possible
to rebuild the parameters of the
network~\cite{tramerStealingMachineLearning2016} or data used during
the training solely from the output of the network~\cite{shokriMembershipInferenceAttacks2017}, which yields concerns
in applications where privacy is paramount, such as healthcare.
The growing interest of industrials on integrating deep neural
networks into their processes, and their use by public institutions
in critical democratic processes (optimization of employement, jury advices,
opinion analysis), demand a paramount level of trust on those programs.

Deep neural networks are composed of layers,
successively computing weighted sums of inputs.
To express non-linear behaviours, they rely
on activation functions, the most popular one being the
rectified linear unit (\relu): $x \rightarrow \max(x,0)$.
This function is
\emph{piecewise-linear}: when the input is strictly negative or positive,
\relu{} acts as a linear function.
As a composition of linear and piecewise-linear functions, the function represented by a neural network is also piecewise-linear.
Regions of the input space that delimit which linear behaviour is taken
by a \relu{} are called \emph{linear regions} or \emph{facets}.
A common idea, stated in~\cite{serraBoundingCountingLinear2018} for instance,
is that the number of facets yielded by a neural network
is a quantification of its expressiveness.
If one would like to explore all possible outputs of a neural network
(for instance, to formally verify a property), one would need to consider
both sides of the \relu{} because of its piecewise linear nature.
A naive exhaustive exploration of the output space will thus rely on
case-splitting, producing cases exponentially in the number of neurons.
This combinatorial explosion is one of the main obstacles to the use
of complete formal verification techniques
(\eg{} Satisfiability Modulo Theory (SMT)
calculus), and must be circumvented before venturing forth.

A recent line of work, however, displayed an interesting idea.
In~\cite{haninDeepReLUNetworks2019},
the authors claim that the number of facets
for networks computing functions from $\mathbb{R}$ to $\mathbb{R}$
is linear in the number of neurons.
In their following work~\cite{haninComplexityLinearRegions2019}, they expand
their results to networks more representative of real-world programs, by
providing an upper bound on the number of facets that is
\emph{not} exponential in the number of neurons but only polynomial;
a shallower bound is present in previous
works on the study of linear regions,
such as in~\cite{serraBoundingCountingLinear2018}.
What if, \emph{empirically}, the number of facets found in trained networks was much lower than the \emph{theoretical} intractable bound on the \emph{maximal} number of facets?
How could we use linear regions to ease
formal verification? Is there a way to reduce the burden of complete
verification tools on deep neural networks? Building up on previous work,
our goal is to address those questions.
If the neural network can be decomposed
into a union of facets, we believe that
verifying a given safety property on each of those
regions will be easier than
-- and still equivalent  to -- verifying
the neural network once on the whole input space. To a lesser extent,
if a considerable number of inputs, say 90\%,
was empirically shown to be in a limited number of regions,
then proving the safety properties on those regions
can be a partial formal verification,
presenting a possibly reasonable trade-off
between cost and exhaustivity.

Our contribution can be summed up by the following:
\begin{enumerate}
    \item we propose an algorithm for
        decomposing an initial verification
        problem into linear subproblems that
        are easier to verify, the
        decomposition and verification being
        embarrassingly parallel,
    \item we provide an in-depth analysis on various
        properties
        of linear regions,
        and we study the influence of techniques reducing
        the number of facets,
    \item we evaluate our approach on different
        verification problems, with linear programming
        and SMT calculus.
\end{enumerate}

\section{Related work}
Over the past years, several lines of work propose different
approaches to formal verification of deep learning programs.
The authors of Reluplex~\cite{katzReluplexEfficientSMT2017},
its successor Marabou~\cite{katzMarabouFrameworkVerification2019}
and the solver
Planet~\cite{ehlersFormalVerificationPieceWise2017}
are the first to aim for exhaustive verification of neural networks,
using SMT calculus.
They propose a reformulation of the simplex algorithm to
lazily evaluate \relu{} and branching heuristics such as
case-splitting on individual
neurons. Their work focus on the algorithmic method
used to solve a non-linear, non-convex problem. Our
technique reformulate the problem as a set of linear
problems to solve, and is independant of the solving
technique used.
Other sound and complete formulations can be found as Mixed Integer Linear
Programming formulation to verify local adversarial robustness, such as
~\cite{tjengEvaluatingRobustnessNeural2017} and branch and
bound~\cite{bunelUnifiedViewPiecewise2017}.

Non-combinatorial approaches also exist.
They generally scale
to wider problems than their exact counterparts, trading for a loss
of precision in the analysis. Symbolic propagation is one of the most
common technique, seen for instance in
~Reluval\cite{wangFormalSecurityAnalysis2018},
CNN-Cert~\cite{boopathyCNNCertEfficientFramework2019}
and ERAN~\cite{singhAbstractDomainCertifying2019}: we
rely on their approaches to propagate information
inside our network as well.
Of course the limit between those two families
is not a clear one:
for example, a combination with MILP formulations to
increase precision can be
seen in~\cite{singhBoostingRobustnessCertification2019}.

Regarding linear regions, a theoretical extension of the
universal approximation theorem applied to robustness
certification was proposed
in~\cite{baaderUniversalApproximationCertified2020}.
An exact enumeration scheme was proposed
by~\cite{serraBoundingCountingLinear2018} using MILP.\@
Our enumeration
scheme closely follow theirs, with some additional
heuristics; we also leverage the obtained linear regions
to perform formal verification, while they do not. They also provide initial
insights by showing a correlation between accuracy and the number of facets.
Using linear regions to increase the
robustness of neural networks had been proposed
in~\cite{croceProvableRobustnessReLU2019}, where the authors describe a
regulation scheme that increases the area of linear
regions, which results in an increase in local robustness
performances. We reimplemented their method and used it in our approach.
Finally, our work is closely related
to~\cite{bakImprovedGeometricPath2020},
where authors propagate linear constraints
within neural networks to
check formal properties on fully-connected
deep neural networks. They use
numerical domains to propagate more information than we
do, namely upper and lower bounds of variables
within each linear regions. They are also able to
overapproximate their propagated set, altough this makes
their method not complete. On the opposite, our
path enumeration is always sound and complete, and only
needs to be called once to verify any property afterward.
To the best of our knowledge, we are the first to propose
an impact analysis of several hyperparameters
on the number of facets for neural networks.

% Some definitions
\section{Background}
\subsection{Activation vectors and facets}

Let $\mathcal{X}$ be a multidimensional input space, subset of
$\mathbb{R}^{Din}$.
Let $\mathcal{Y}$ be an output space, typically a subset of $\mathbb{R}^{Dout}$.
Let $f$ be a trained neural network of $L$ layers,
computing values from $\mathcal{X}$
to $\mathcal{Y}$: $f: \mathcal{X} \rightarrow \mathcal{Y}$.
Each layer computes a multidimensional input and produces a
multidimensional output, both represented as multidimensional arrays (also known
as tensors). Each cell of a tensor is called a neuron.
A layer $l_{i}$ has an input in $\mathbb{R}^{D_{(i-1)}}$ and an output in
$\mathbb{R}^{D_i}$, for $i=2..L$, with $D_1 = Din$ and $D_L = Dout$. In the
rest of this paper, we will denote a layer by $l$ to avoid cluttering.

We consider here a network for which each layer $l$
is composed
of a linear application, followed by a \relu{}
activation function on all the resulting neurons.
Parameter tensors are obtained after training and do not change while using
the resulting program: they are used in the various mathematical operations
occuring during the layers computations.
The only variables are the vectors in $\mathcal{X}$.
For a given multidimensional input
$\vec{x} \in \mathcal{X}$, each neuron of the layer $l$
can be either \emph{active},
if their value before the application of
\relu{} is greater than 0,
or \emph{inactive} when this value is stricly lower than 0.
We denote by $\state_{facet}^{l}$ the activation state of \relu{} neurons for
a given layer $l$: an active neuron is denoted by $1$, an inactive neuron
by $0$. As an example, for the network in \cref{fig:nn_example},
$\state^{1}_{\facet} = \left( 0, 0, 1\right)$.

We call a \emph{facet} the subset $\facet$
of the input space generating a certain activation pattern
$\state^{l}_{\facet}$. The network yields the same activation pattern
for all inputs within this region. Such a facet describes a linear region,
because all \relu{} have a fixed behaviour within it; thus the network
with inputs reduced to $\facet$ is simply a composition of linear applications.

\subsection{Building facets}\label{subsec:building_facets}
\begin{figure}[h!]
    \includegraphics[width=\textwidth]{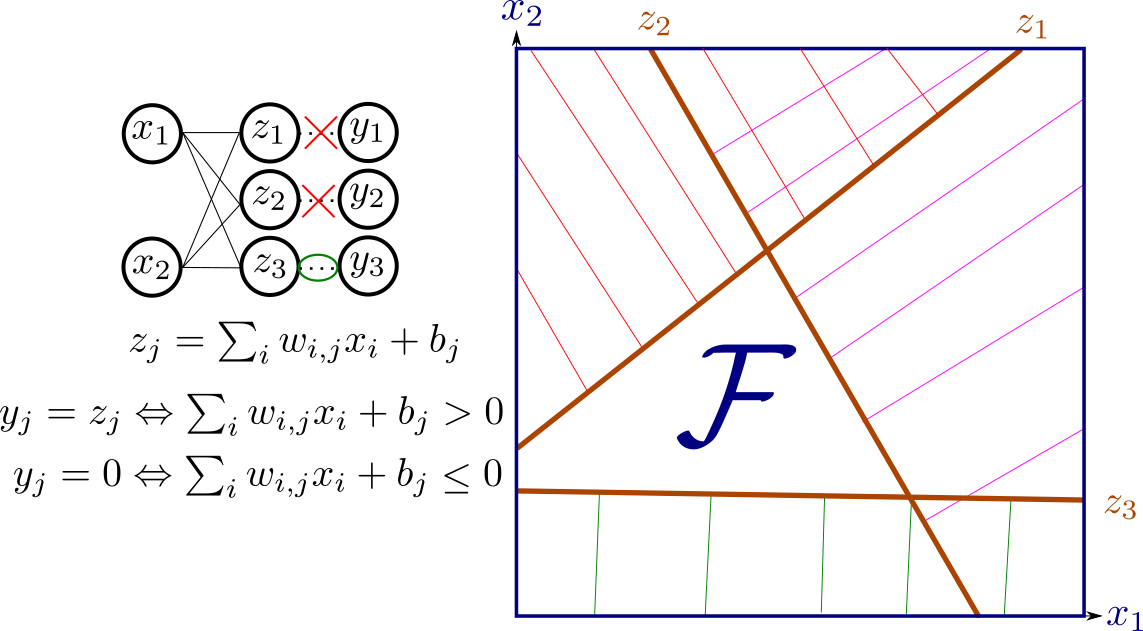}
    \caption{A toy network where each $y_{i}$ is a \relu node, 
    $x_{i}$ real values, and $w_{i,j}$ parameters. The right
    figure shows the facets of the input space. Each activation of each \relu
    node induces a half-space. Dashed zones depict half-spaces unreachable
    by the activation states of $y_{i}$. $\facet{}$ is the intersection of
    half-spaces induced when $y_{3}$ is active and $y_{1}$ and $y_{2}$
    are inactive.
    }\label{fig:nn_example}
\end{figure}

Let $n_{i,l}$ be a neuron at layer $l$. If this neuron is active,
it means that the lower bound of its input is non-negative.
Since the value of this neuron is the result
of previous affine transformations,
it follows that being activated can be expressed as a linear constraint for its predecessors.
For example, if
the affine transformation in layer $l$ is a matrix multiplication
of elements $w_{i,j}^l$ with
outputs $y_{j,(l-1)}$ of the previous layer,
the linear constraint is
\begin{equation}
    n_{i,l} = \sum_{j}w_{i,j}^l\,y_{j,(l-1)}\geq 0\label{eq:active}
\end{equation}
Similarly, an inactive neuron yields the constraint
\begin{equation}
    n_{i,l} = \sum_{j}w_{i,j}^l\,y_{j,(l-1)} <  0\label{eq:inactive}
\end{equation}
As for a given activation pattern the inputs $y_{j,l-1}$ of layer $l$ are affine functions of the input $x$ of the network, such constraints can be expressed in terms of hyperplanes in the input space.
Each neuron generates one such constraint in the input space; a facet is thus
the conjunction of those constraints from all neurons together.
Geometrically,
a facet can be seen as the convex polytope described by the set of constraints
resulting from the activation pattern.
See \cref{fig:nn_example} for an illustration of facets on a toy network.

In the rest of this paper, we aim to \emph{formally verify} a neural
network: given a network $f$, a precondition on the input space
$\mathcal{D} \subset \mathcal{X}$
and a postcondition on the output space $\mathcal{P} \subset \mathcal{Y}$,
we want to provably ensure that
\[
\forall x \in \mathcal{D} \rightarrow f(x) \in \mathcal{P}
\]
The form of the pre and postcondition vary according to the property we want
to check. For instance, local adversarial robustness around a sample
would be expressed as, given $x \in \mathcal{X}, \forall \varepsilon <
\varepsilon_{0}, f(x+\varepsilon) = f(x)$. For safety properties of the ACAS
benchmark described for instance in \cite{manfrediIntroductionACASXu2016},
the precondition on the inputs and outputs are linear constraints.

% Contrib: algorithm and implementation of facet finding
\section{Divide and conquer on linear regions}
Linear operations are easier to verify than networks
with \relu{}, since
they do not produce case splits on solvers.
If we somehow have an exhaustive list of \emph{actually reached} facet
for our problem at hand, it would be possible to verify each facet
independently.
Even if the theoretical number of facet is exponential
in the number of neurons,
a network does not actually exploit the whole set of possible linear regions.
For a simple task, a deep and wide network seems to only use a
small partition of the total input space.
Facets can also have a wider support in the input space, which may indicate
that this particular subset of input is much more relevant for the problem at
hand.
On the other hand, we want to perform a sound and
complete verification. Sound means that if our method
answers that a system is safe, then it is actually safe;
complete means that if a faulty behaviour exists for
our problem, it will be spotted by our procedure.
The key point is thus to exhibit a procedure to enumerate all the facets that
are actually within the (constrained) input space, while excluding
facets that, while theoretically possibly expressed in our network, are not
present. In other words, we want to find all
$\facet_{i}$ such that
$\bigcup_{i}\facet_{i} = \mathcal{X}$.

\subsection{Enumeration of facets}
Our approach is to start from the beginning of the network
and proceed neuron by neuron. Using an initial bounding
box $\mathcal{D}$ as an initial constraint on the inputs,
we iteratively build the
linear constraints composing the neural network,
as described in \cref{subsec:building_facets}.
Linear operations are directly written as linear equalities
in a stack $s$.
When a \relu{} neuron $y_{i}$ is considered, the algorithm solves
a problem consisting on the conjunction of the constraints
in $s$ and the linear constraints describing the
activation pattern of $y_{i}$.
The active (resp.\ inactive) pattern
yield the constraint described by \cref{eq:active},
(resp.\ \cref{eq:inactive}).
If only one of the two activation state is possible, then
the constraints describing this state are added
to $s$, and the algorithm goes through the
next neuron.
If both activations are possibles, then the problem stack
is copied. Active constraints are added to the first
copy, while inactive constraints are added to the second
one.
Since the two sub-problems are independants, this algorithm
can be parallelized. See \cref{alg:enum_facets} for a pseudo-code description.

\begin{algorithm}[h!]
    \SetKwFunction{BuildExpression}{BuildExpression}
    \SetKwFunction{BuildConstraints}{BuildConstraints}
    \SetKwFunction{Solve}{Solve}
    \SetKwFunction{Length}{Length}
    \SetKwFunction{SendToNP}{SendToNewProcess}
    \SetKwFunction{Range}{Range}
    \SetKwData{stack}{stack}
    \SetKwData{facets}{facets}
    \SetKwData{stackcp}{stack\_copy}
    \SetKwData{linexprs}{lin\_exprs}
    \SetKwData{reluneurons}{relu\_neurons}
    \SetKwData{constraint}{constraint}
    \SetKwData{net}{$\mathcal{N}$}
    \SetKwData{neuron}{neuron}
    \SetKwData{act}{active\_expr}
    \SetKwData{ina}{inactive\_expr}
    \SetKwData{i}{index}
    \SetKwData{len}{len}
    \SetAlgoLined
    \KwData{An input space domain $\mathcal{D}$,
    a list of all neurons in the network \net}
    \KwResult{A set of linear problems describing all
    feasible facets for the input space}
    \tcp{build the expressions for each neurons}
    \linexprs, \reluneurons =
    \BuildExpression{\net}\;
    \len = \Length{\reluneurons}\;
    \i = 0\;
    \stack = $\mathcal{D}$\;
    \tcp{a shared resource between processes}
    \facets = $\emptyset$\;
    \While{\i  < \len }{
        \tcp{linear expressions describing
        the activation state for a given neuron}
        \act, \ina = \BuildConstraints{\neuron} \;
        \tcp{propagate only the feasible activations}
        \If{\Solve{\stack $\bigcap$ \act}}{
            \If{\Solve{\stack $\bigcap$ \ina}}{
                \stackcp = \stack.copy()\;
                \stackcp.push(\act) \;
                \tcp{send the copied stack to a new
                instance of the algorithm}
                \SendToNP{\stackcp, \i + 1}\;
                \tcp{proceed in the current process
                with the other possible state}
                \stack.push(\ina) \;
            }
            \Else{
                \stack.push(\act) \;
            }
        }
        \Else{
            \stack.push(\ina) \;
        }
        \i = \i + 1 \;
    }
    \tcp{when all neurons have been analyzed,
        add the resulting linear constraints to the
    list of facets}
    \facets.append(\stack)\;
    \Return{\facets}\;
        \caption{Counting facets}\label{alg:enum_facets}
    \end{algorithm}

Once we obtain the set of all relevant facets, it is possible to build the
corresponding linear functions. This set of linear functions represent
all the possible behaviours of the network on its input space.
Verification of the property can then be launched on each linear function;
since they are independant problems: parallelization can also be used.
More formally, let us consider a facet set
$\bigcup_{i}\facet_{i}$ for a network $f$, an input space $\mathcal{X}$, an
output space $\mathcal{Y}$, a precondition on the input space
$\mathcal{D} \subset \mathcal{X}$ and a postcondition on the output
space $\mathcal{P} \subset \mathcal{Y}$. We aim to formally
verify that $x \in \mathcal{D} \implies f(x) \in \mathcal{P}$.
Partitionning consists on adding to the network's control flow
the constraint on the inputs yielded by $\facet_{i}$, and to force
the corresponding activation state for all \relu{} neurons.
The resulting function $\restr{f}{\facet_{i}}$ is thus a composition of
linear operations: original matrix multiplications
and active or inactive \relu{} (which are diagonal matrices multiplied to the
pre-activation inputs). Then, the verification problem becomes
$x \in \mathcal{D} \cap \facet_{i} => f(x) \in \mathcal{P}$.
We Divide the Input Space into COnvex polytopes, thus we will be referencing
our technique as DISCO in the rest of this paper.

\subsection{Evaluation}

We implemented DISCO in OCaml, within the tool
Inter Standard Artificial Intelligence Encoding Hub
%(ISAIEH)\footnote{\url{https://git.frama-c.com/pub/isaieh}}.
(ISAIEH)\footnote{\url{https://git.frama-c.com/pub/isaieh}, only contains the
SMT implementation; full DISCO implementation is under review for open source}.
ISAIEH leverages the ONNX standard neural network format
to formulate an intermediate representation.
This intermediate representation can then be compiled
down to a standard SMT formula representation,
SMTLIB~\cite{barrettSMTLIBStandard2017},
or to linear programming problems, using the
formulation proposed
in~\cite{tjengEvaluatingRobustnessNeural2017}.
It can also be manipulated to apply various simplification
technique.
ISAIEH performs symbolic propagation to compute
the hyperplanes delimiting
facets boundaries during a forward pass,
the building of facets
is then made according to~\cref{alg:enum_facets}.
The intermediate representation is an acyclic
directed graph $(V,E)$ where vertices $V$ represent
computations of deep learning techniques
(matrix multiplication, convolution, pooling),
linked by edges $E$ to
represent the flow of calculus within the network.
Each $V$ describes the
input, output, operation occuring as well as some
parameters if necessary.
This graph is then computed by an output formatter that
rewrites the control flow under the SMTLIB/LP format;
multidimensional operations are rewritten to be
compatible with several SMT theories and LP formulations.
Supported operations is a subset of ONNX standard operators
\footnote{\url{https://github.com/onnx/onnx/blob/master/docs/Operators.md}}.
The linear programming implementation was made with
the Python programming language, and Gurobi~\cite{gurobi}
was used as a LP solver (version 9.1.1).
For the SMT verification,
z3~\cite{demouraZ3EfficientSMT2008} was used
(version 4.8.10).

We consider two synthetic, easy to analyze problems:
\begin{enumerate}
    \item multiplication between $N$ floating points
        numbers sampled between 0.5 and 2; this
        problem will be called
        called \emph{N-multiplication} in
        the rest of the paper
    \item detection of the presence of
        an obstacle within a given area; this
        problem will be called
        \emph{N-perception} in the rest of the paper
\end{enumerate}

For those problems, we study different architectures.
All of them are fully-connected networks.
$N-multiplication$ networks have three hidden layers,
$N-perception$ ones have two hidden layers.
 Details are on \cref{tab:arch}.
\begin{table}
    \centering
    \begin{tabular}{C{4cm}|C{2cm}|C{2cm}|C{2cm}}
        name & $L_1$ & $L_2$ & $L_3$ \\
        \hline
        simple & $N\times 2$ & $N$ & $N/2$ \\
        big & $N\times3$& $N$ & $N/2$ \\
        super & $N\times4$& $N\times2$ & $N$ \\
        perception & $N/2$ & $N/4$ & -- \\
    \end{tabular}
    \caption{Number of neurons for the different architectures.
        $N$ denotes the dimension of the input,
    $L_{i}$ the $i-th$ layer of the network}\label{tab:arch}
\end{table}

For each of the two problems,
we aim to count the number of facets,
then verify if the network repect its specification.
For $N-multiplication$, we check if the network
can indeed produce multiplication results within
the tolerance. As formulating this problem directly
is impossible due to linear programming limitations,
we instead check the if following property is verified:
\begin{equation}
    \sum_{k=1}^N x_k + 1 - \frac{5}{4} N + \alpha_N \;\;\; \leqslant \;\;\;\prod_{k=1}^N x_k
    \label{eq:multiplication_property}
\end{equation}
with $\alpha_N = 0$ if the input dimension $N$ is even and $\alpha_N = \frac{1}{4}$ otherwise. This
property is always true for our input space $[0.5, 2]^N$. A proof of this inequality
can be found in the appendix.
For $N-perception$, we check the following two properties:
\begin{enumerate}
    \item if an input with at least one obstacle
        (modeled as white pixel) in the lower half of
        the image is presented to the network, the output
        will always be over 0
    \item if an input with no obstacle on
        the lower half of the image is presented
        to the network, the output will always be below 0
\end{enumerate}
Experiments were done on a Dell Precision 5530 with
an Intel Core i7-8850H CPU, 2.6Ghz,
and Ubuntu 20.04.1 LTS as operating system.
See \cref{tab:results} for partial results, full
runtimes in the appendix.
For each network,
the first column describes the runtime
of verification without rewritting, while the second column
describes the runtime of verification for our rewritting technique. To be fair,
the runtime of the enumeration scheme
is also noted on the third column. Solving with DISCO or with standard
MILP formulation always returns the same result.
Note however that the splitting in linear
regions is independant from the
verification problem: costly enumeration algorithms
could be used to obtain
the facets of a neural network once,
then verification could happen afterward. Also, classical
MILP formulation returns a failure immediately, while our current
implementation of DISCO waits for the result of verification for
all facets to finish before returning a result: returning a failure
immediately would decrease the runtime of the verification part
(preliminary experiments on networks with a high number of facets show that
    failures are detected early: guiding the search with a fail-first heuristic
would prove useful).
Chosen networks are those with the maximum accuracy,
with similar architectures. $N-multiplication$ problems
were solved using Linear Programming, while $N-perception$
problems were solved using SMT, QF\_LRA theory.
We note that the speed-up for the problem verification
is much higher with SMT than LP.
A possible explanation is that the number of facets with
$N-multiplication$ being much lower than
in $N-percetion$, the additional cost
of counting and parallelizing verification on each facet
is not worth.

\begin{table}
    \centering
    \begin{tabular}[h!]{C{3cm}|C{3cm}|C{3cm} C{3cm}|C{2cm}}
        Dimension of input & No split & DISCO verification
        & Facet enumeration & Total time DISCO \\
        \hline
        3 super & 0.769s$\pm$0.0205 & \textbf{0.145s}$\pm$0.012 & 2.69s$\pm$0.0596 & 2.83s \\ 
        3 super mmr & 0.498s$\pm$0.00295 & \textbf{0.184s}$\pm$0.0142 & 1.86s$\pm$0.0142 & 2.05s \\ 
        4 big & 0.25s$\pm$0.00423 & \textbf{0.0972s}$\pm$0.00764 & 0.663s$\pm$0.0156 & 0.76s \\ 
        4 big mmr & \textbf{0.454s}$\pm$0.0104 & 1.43s$\pm$0.0444 & 16.9s$\pm$0.0931 & 18.3s \\ 
        4 super & 5.43s$\pm$0.31 & \textbf{0.71s}$\pm$0.0591 & 13.1s$\pm$0.859 & 13.8s \\ 
        4 super mmr & 3.69s$\pm$0.133 & \textbf{2.77s}$\pm$0.174 & 35.7s$\pm$1.41 & 38.4s \\ 
        5 simple & \textbf{0.0179s}$\pm$0.00596 & 0.0771s$\pm$0.0077 & 0.699s$\pm$0.0124 & 0.776s \\ 
        5 simple mmr & \textbf{0.0204s}$\pm$0.00084 & 0.346s$\pm$0.0174 & 3.75s$\pm$0.0581 & 4.09s \\ 
        5 big & \textbf{0.0279s}$\pm$0.00148 & 1.31s$\pm$0.0622 & 17.4s$\pm$0.283 & 18.7s \\ 
        5 big mmr & \textbf{0.0154s}$\pm$0.000531 & 1.48s$\pm$0.0513 & 18.8s$\pm$0.0867 & 20.3s \\ 
        6 simple & \textbf{0.0264s}$\pm$0.00124 & 0.988s$\pm$0.0693 & 11.6s$\pm$0.186 & 12.6s \\ 
        6 simple mmr & \textbf{0.0291s}$\pm$0.00132 & 1.3s$\pm$0.0342 & 16s$\pm$0.149 & 17.3s \\ 
        7 simple & \textbf{0.0474s}$\pm$0.00158 & 16.8s$\pm$0.831 & 227s$\pm$8.51 & 244s \\ 
        7 simple mmr & \textbf{0.0306s}$\pm$0.0016 & 1.09s$\pm$0.0348 & 15.6s$\pm$0.555 & 16.7s \\ 
        8 simple & \textbf{0.0484s}$\pm$0.00551 & 1.65s$\pm$0.113 & 27.2s$\pm$0.576 & 28.8s \\ 
        8 simple mmr & \textbf{0.12s}$\pm$0.00269 & 1.72s$\pm$0.0988 & 28.9s$\pm$0.697 & 30.6s \\ 
        \hline
        \hline
        $5\times5$ perception & 132s & 23.7s & 0.86s & \textbf{24.56s} \\
        $7\times7$ perception & TIMEOUT & 1393s &  15.38s & \textbf{1406.38s}\\
    \end{tabular}
    \caption{Runtime for different problems. TIMEOUT is set at 10000s. Figures
        are mean taken over 10 runs, standard deviation is reported next to the
    $\pm$ symbol}\label{tab:results}
\end{table}

\subsection{Further reducing the number of facets using maximum margin
regularization}
Formally proving a property using DISCO
require to enumerate
all possibly achievable facets.
Even if their practical number is far below
theoretical upper bounds, any existing method
reducing it is worth studying.
Such a method exists: maximum margin regularization (MMR), presented
in~\cite{croceProvableRobustnessReLU2019}.
The authors propose to modify the learning objective of the neural
network to maximize the distance between a sample and nearby facets boundaries.
Neural networks tend to ``push away'' the boundaries, resulting on fewer
facets for a fixed $\mathcal{X}$.
More formally, let us consider a facet $\facet_{i}$. This facet
is neighbored by $k$ others, leading to $k$ boundaries.
Each of those boundaries
are hyperplanes yielded by $\facet_{i}$ and its neighbours, their equation
can then be written as $V^k_{\facet_{i}}$.
Here, $V^k_{\facet_{i}}$ is the orthogonal vector to the hyperplane
constituting the $k-th$ boundary with $\facet_{i}$.
For any sample $s$ within $\facet_{i}$, the distance between $s$ and
a hyperplane defined by $V^k_{\facet_{i}}$
is $\langle V^k_{\facet_{i}},s\rangle$ (where
$\langle \cdot,\cdot \rangle$ denotes the scalar product).
In their paper, they compute this distance
and aim to maximize it. Another
distance towards decision boundaries
is also computed, but since we focus on
regression tasks, the notion of decision
boundaries is not relevant here. The
final term added in the cost function of the network is
then, with $\gamma_{rb}$ a parameter and $p$ either 1, 2 or $\infty$:
\begin{equation}
    \max{(0, 1 -
            \min_{k}(\frac{
        \langle V^k_{\facet_{i}},s\rangle +
    }{\norm{V^k_{\facet_{i}}}{p}} *
\frac{1}{\gamma_{rb}}))}
    \label{eq:mmr}
\end{equation}

We reimplemented their method and applied
DISCO on networks trained with
MMR, for $N-multiplication$ problems.
Results are available \cref{tab:results}
and \cref{fig:num_facets}.
First, the effective number of facets is reduced
with MMR training, leading to a reduction
of one or two order of magnitudes in certain cases, which
leads to lower verification times.
However, we note that training a network with
MMR has an impact on accuracy.
Achieving comparable performance while reducing
the number of facets is
a difficult task. This is no surprise, since a high
number of facets denotes a wide variety of possible
behaviours for the network: reducing
the number of facets means the network's
behaviour will be less complex.
It is then necessary to find a tradeoff between
accuracy and robustness;
tradeoff that may be sometimes very difficult
to achieve~\cite{tsiprasRobustnessMayBe2018}.

% Contrib: Analysis of facets
\section{Studies on facets}
So far, we presented a methodology to use facets to ease formal verification.
Some characteristics of those facets remain however unknown. What is the
volume occupied by a facet on the input space? Are all facets activated
uniformly? Which parameters influence the number of facets?
In this contribution, we perform an analysis of the facets of our networks.

\subsection{Towards counting facets and beyond}
\begin{figure}[h!]
    \centering
    \includegraphics[width=\textwidth]{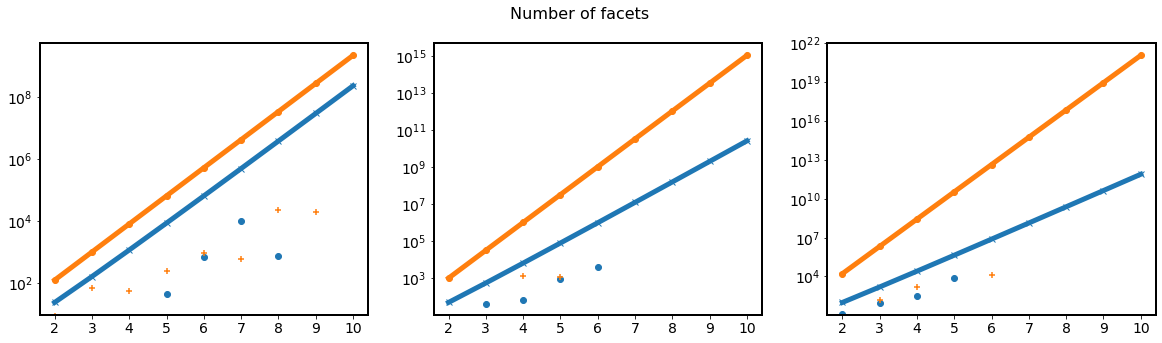}
    \caption{Number of facets for our three architectures. x-axis is the
        dimension of the input, y-axis is the number of facets. Upper line with
        dots is the naive, $2^n$ bound. Middle line with crosses is the
    bound proposed by \cite{haninDeepReLUNetworks2019}. Individual dots
are the best performing networks for our experiments:
dots are trained normally, crosses are trained with MMR.
y-scale is logarithmic}\label{fig:num_facets}
\end{figure}

The initial motivation of this work was that the theoretical number of facets
was far over the actual number, and that it was possible to leverage
facets for formal verification. For our problem at least, this seems to
be the case. We took the best performing network trained both with
and without MMR. Even without MMR, most of the networks are about one or
two order of magnitude below the bound proposed
in~\cite{haninDeepReLUNetworks2019}. The progression still seems to
be exponential in the number of neurons however, so this remains a hard
problem. See \cref{fig:num_facets} for more details.

\subsection{Not all facets are equals}
Reducing the number of facets is a way to reduce the complexity of
verification. When starting the verification, the solver will try each
facet without priorizing one over the other. This relies on the assumption
that all facets are activated relatively evenly, that is to say, that
each achievable facet has an equal chance to be activated by an input point.
If some facets were more frequent than others, a possible approach would be
to identify the most used facets and prioritize verification on those.
Also, the frequency of a facet's occurence can be a good proxy to estimate
the space occupied by the facet in the input space.

\begin{figure}[h!]
    \centering
    \begin{minipage}[h!]{0.44\textwidth}
        \includegraphics[width=\textwidth]{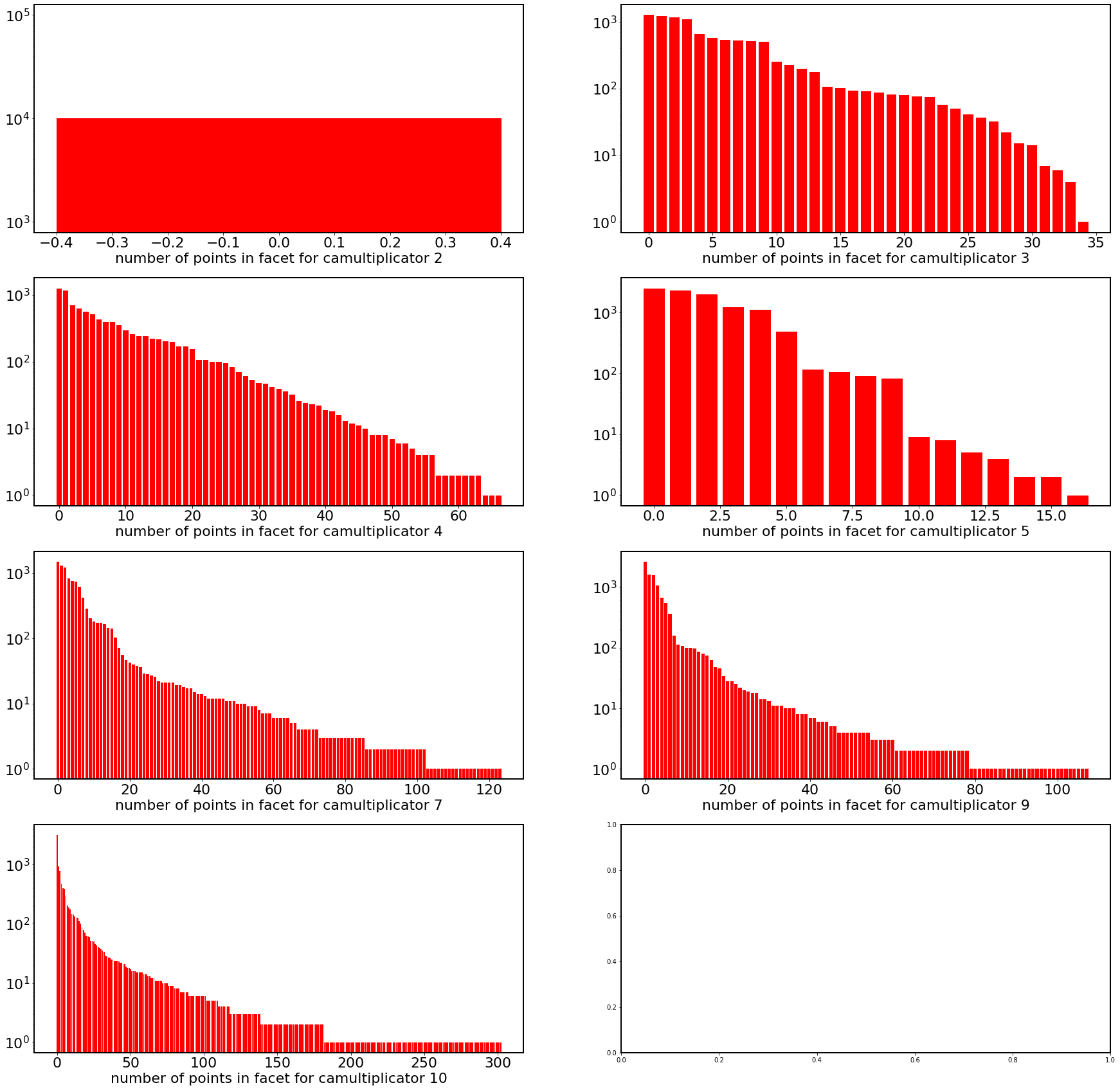}
    \end{minipage}
    \begin{minipage}[h!]{0.44\textwidth}
        \includegraphics[width=\textwidth]{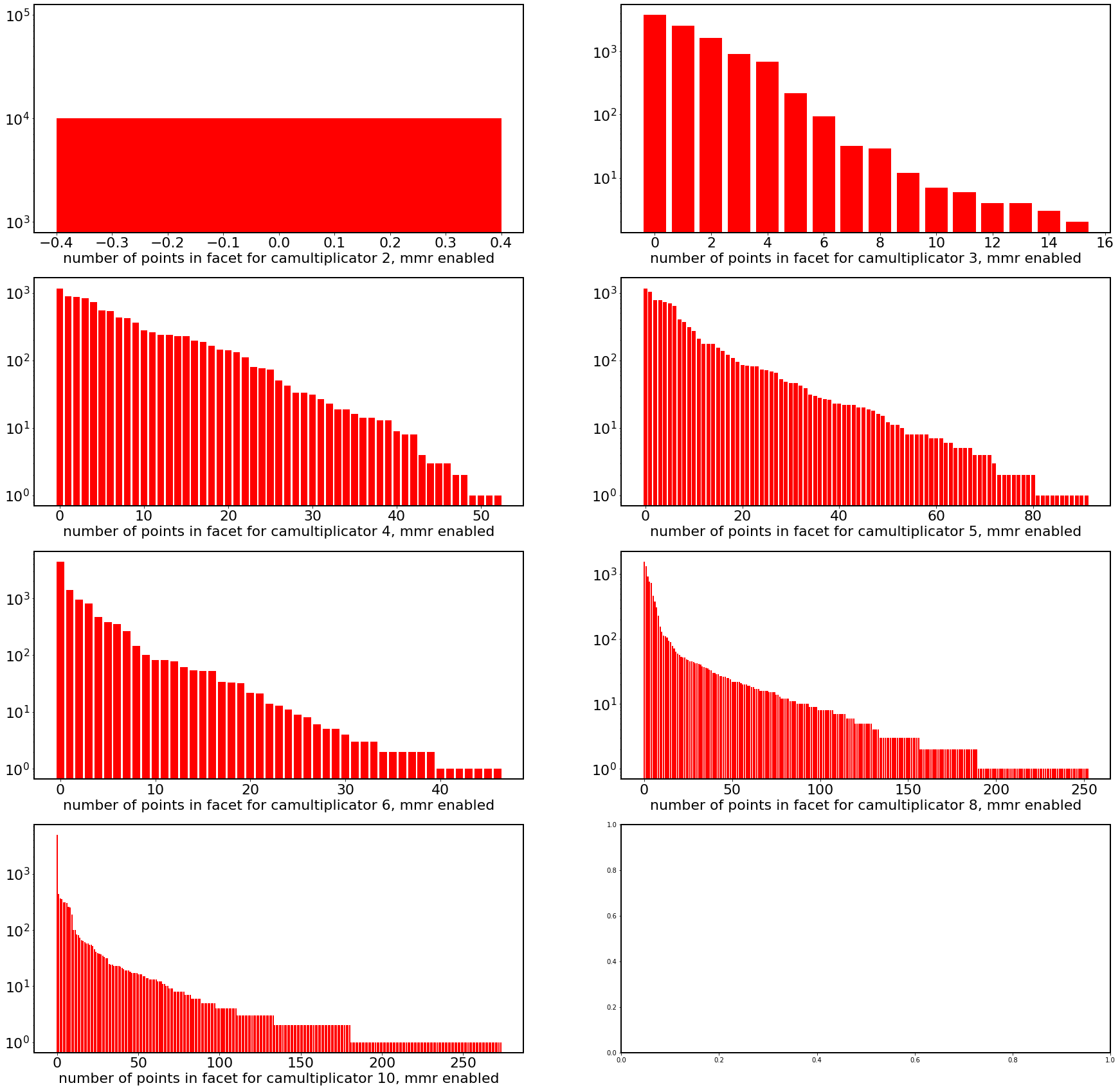}
    \end{minipage}
    \caption{For 10000 random samples, x-axis denotes an unique facet, y-axis
        denotes the number of points that activated this specific facet.
        Scale is logarithmic}\label{fig:sample}
\end{figure}

We performed uniform sampling on selected networks
(on the same distribution of the training set),
and collected the number of points contained in each
facet. Some results are available
on \cref{fig:sample}. We note that for
some programs, a very small number of facets are
concentrating almost 70\% of the possible inputs.

\subsection{What makes facets shine?}
\begin{figure}[!h]
    \centering
    \includegraphics[width=\textwidth]{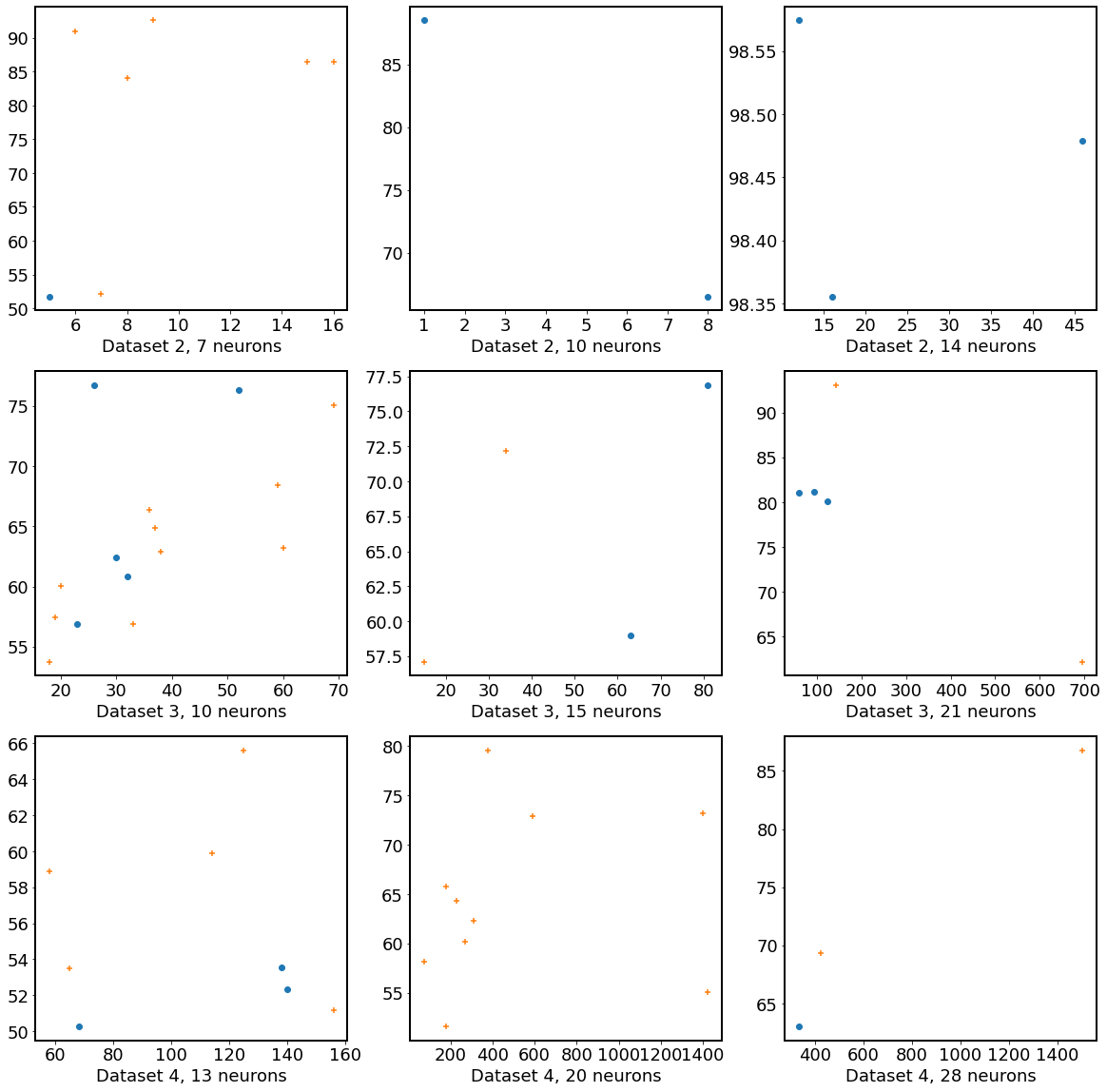}
    \caption{Graph summing up the performances of
        several networks. x-coordinate denotes the
        number of facets, y-coordinate the
    accuracy of the network}\label{fig:mmr}
\end{figure}

Apart from using an explicitly designed training scheme to reduce the
number of facets, other parameters may influence this number and
its growth (or decrease) during training and after.
We present on \cref{fig:mmr} a summary of all the experiments we made,
for different parameters. Obviously, networks trained with MMR do have less
facets than the others. Among parameters we changed are the neural network
starting learning rate, the training time, initialization seed and
parameters related to MMR\@:
$\gamma_{rb}$ and the loss used for distance calculation.
Using $l_{1}$ and $l_{\infty}$ norms tend to slightly increase the number
of facets for the same accuracy. Interestingly, $l_{2}$ norms provide about the
same accuracy but with lesser facets; a higher $\gamma_{rb}$ results in lower
facets for similar accuracy. This may come from the low complexity of the
function we are studying on the input space (multiplication of two real values
on $[0.5,2.0]$ is a saddle with very low slopes).

% Discussions and possible extensions
\section{Discussion and perspectives}

We presented a method of partitionning for the input
space into linear subregions, or facets.
We used classical linear programming solvers to enumerate facets and
launch verification on those facets.
Our problem (regression) is specific and the size of our networks
is relatively small. Further research is necessary to assess the usefulness
of our technique on high-dimensional input networks, on classification tasks.

The enumeration of facets is a pre-requisite for our
method to work; with a lot of case splits, even with
parallelism, it remains a bottleneck and computationnaly
expensive. The tested version only implements
basic heuristics; more elaborated techniques used elsewhere in the
literature, for instance overapproximations or using
pre-calculated bounds, could certainly improve our method.
Using a training scheme forcing the network to reduce the number of facets
showed encouraging results, with
sometimes a reduction of several order of magnitude in the number of facets.
However, this regularization technique comes with a tradeoff with
accuracy, and requires to train a network from scratch. One could adapt
such regularizer to avoid retraining entirely the network. The tradeoff
between robustness and accuracy is not unique to our method, as almost
all techniques in the literature face this ``No Free Lunch'' situation.

Another possible improvement would be in the facets
themselves. Indeed, the number of linear regions stays
high with very deep neural networks, limiting the gain of
parallelism.
To do so, one could devise a merging scheme between facets, in order
to reduce the number of actual facets while preserving
the neural network expressivity. This would also lead to
a modification of the networks behaviour that should be
carefully controlled.

The non-uniform repartition of points within facets is of high interest.
Even if we are not able to prove the whole set of reachable facets for a given
network, being able to identify which facets concentrate most points is a
precious information for formal tools, allowing them to guide the verification
process towards most sensitive points. A fail-first heuristic search would
certainly benefit from this guidance.

\printbibliography
\newpage
\appendix
\section{Full runtime results for various problems}

\begin{table}
    \centering
    \begin{tabular}[h!]{C{3cm}|C{3cm}|C{3cm} C{3cm}|C{3cm}}
        Dimension of input & No split & DISCO verification
        & Facet enumeration & Total time DISCO \\
        \hline
        2 simple mmr & \textbf{0.0034s}$\pm$0.00065 & 0.0333s$\pm$0.00147 & 0.0327s$\pm$0.00166 & 0.066s \\ 
        2 big & \textbf{0.00116s}$\pm$0.000321 & 0.0245s$\pm$0.00135 & 0.000648s$\pm$8.31e-05 & 0.0251s \\ 
        2 super & 0.17s$\pm$0.00795 & \textbf{0.0444s}$\pm$0.0157 & 0.219s$\pm$0.00982 & 0.263s \\ 
        3 simple mmr & \textbf{0.00514s}$\pm$0.000533 & 0.0793s$\pm$0.00836 & 0.393s$\pm$0.014 & 0.472s \\ 
        3 big & \textbf{0.0413s}$\pm$0.00186 & 0.0615s$\pm$0.00758 & 0.321s$\pm$0.0126 & 0.383s \\ 
        3 super & 0.769s$\pm$0.0205 & \textbf{0.145s}$\pm$0.012 & 2.69s$\pm$0.0596 & 2.83s \\ 
        3 super mmr & 0.498s$\pm$0.00295 & \textbf{0.184s}$\pm$0.0142 & 1.86s$\pm$0.0142 & 2.05s \\ 
        4 simple mmr & 0.244s$\pm$0.00284 & \textbf{0.0799s}$\pm$0.0114 & 0.57s$\pm$0.0108 & 0.65s \\ 
        4 big & 0.25s$\pm$0.00423 & \textbf{0.0972s}$\pm$0.00764 & 0.663s$\pm$0.0156 & 0.76s \\ 
        4 big mmr & \textbf{0.454s}$\pm$0.0104 & 1.43s$\pm$0.0444 & 16.9s$\pm$0.0931 & 18.3s \\ 
        4 super & 5.43s$\pm$0.31 & \textbf{0.71s}$\pm$0.0591 & 13.1s$\pm$0.859 & 13.8s \\ 
        4 super mmr & 3.69s$\pm$0.133 & \textbf{2.77s}$\pm$0.174 & 35.7s$\pm$1.41 & 38.4s \\ 
        5 simple & \textbf{0.0179s}$\pm$0.00596 & 0.0771s$\pm$0.0077 & 0.699s$\pm$0.0124 & 0.776s \\ 
        5 simple mmr & \textbf{0.0204s}$\pm$0.00084 & 0.346s$\pm$0.0174 & 3.75s$\pm$0.0581 & 4.09s \\ 
        5 big & \textbf{0.0279s}$\pm$0.00148 & 1.31s$\pm$0.0622 & 17.4s$\pm$0.283 & 18.7s \\ 
        5 big mmr & \textbf{0.0154s}$\pm$0.000531 & 1.48s$\pm$0.0513 & 18.8s$\pm$0.0867 & 20.3s \\ 
        5 super & \textbf{0.102s}$\pm$0.003 & 16.2s$\pm$0.864 & 381s$\pm$11.6 & 398s \\ 
        6 simple & \textbf{0.0264s}$\pm$0.00124 & 0.988s$\pm$0.0693 & 11.6s$\pm$0.186 & 12.6s \\ 
        6 simple mmr & \textbf{0.0291s}$\pm$0.00132 & 1.3s$\pm$0.0342 & 16s$\pm$0.149 & 17.3s \\ 
        6 big & \textbf{0.0428s}$\pm$0.00292 & 6.94s$\pm$0.249 & 90s$\pm$2.01 & 96.9s \\ 
        6 super mmr & \textbf{0.201}s$\pm$0.038 & 44.1s$\pm$7.24 & 576s$\pm$62.8 & 620s \\ 
        7 simple & \textbf{0.0474s}$\pm$0.00158 & 16.8s$\pm$0.831 & 227s$\pm$8.51 & 244s \\ 
        7 simple mmr & \textbf{0.0306s}$\pm$0.0016 & 1.09s$\pm$0.0348 & 15.6s$\pm$0.555 & 16.7s \\ 
        8 simple & \textbf{0.0484s}$\pm$0.00551 & 1.65s$\pm$0.113 & 27.2s$\pm$0.576 & 28.8s \\ 
        8 simple mmr & \textbf{0.12s}$\pm$0.00269 & 1.72s$\pm$0.0988 & 28.9s$\pm$0.697 & 30.6s \\ 
        \hline
        \hline
        $5\times5$ perception & 132s & 23.7s & 0.86s & \textbf{24.56s} \\
        $7\times7$ perception & TIMEOUT & 1393s &  15.38s & \textbf{1406.38s}\\
    \end{tabular}
\end{table}

\section{Full grid of experiments for $N-multiplication$}

\begin{figure}[!ht]
    \centering
    \includegraphics[width=0.7\textwidth]{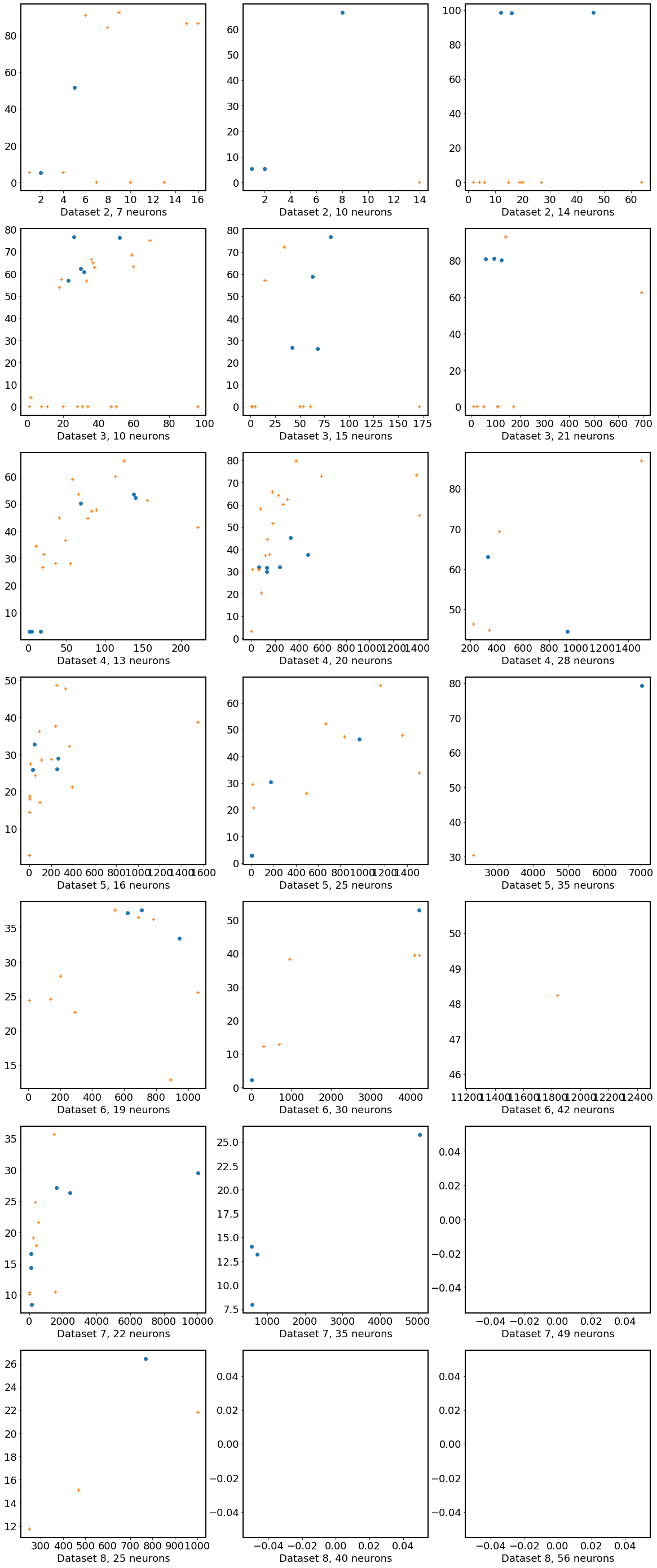}
    \caption{x-axis: number of facets, y-axis: accuracy}\label{fig:annx}
\end{figure}

See \cref{fig:annx}.

\section{Proof for \cref{eq:multiplication_property}}

Though $f: x \mapsto x^n$ is convex for any $n \in \mathbb{N}$, the multiplication of $n$ variables $f: (x_1, x_2, \dots, x_n) \mapsto \prod_k x_k$ is not convex.

For instance for $n=2$, the surface $f: x,y \mapsto xy$ is a saddle surface

\subsection*{Formulation}

We aim at finding a linear (affine) lower bound and a linear upper bound to the multiplication $\prod_{k=1}^n x_k$  of $n$ variables $x_k$ in $[0.5, 2]$.

\subsection*{Upper bound}

First, note this inequality between the product and the average:
$$\prod_{k=1}^n x_k \;\;\leqslant\;\; \left( \frac{\sum_{k=1}^n x_k}{n} \right)^n$$

\subsubsection*{Proof}
$\log$ is concave; consequently, the average of logs is smaller than the log of the average:
$$ \sum_{k=1}^n \frac{1}{n} \log(x_k) \;\; \leqslant \;\; \log \left( \frac{\sum_{k=1}^n x_k}{n} \right)$$
hence
$$ \sum_{k=1}^n \log(x_k) \;\; \leqslant \;\; n \log \left( \frac{\sum_{k=1}^n x_k}{n} \right)$$
and taking the exponential we get the desired result. Note that we use the positivity of all $x_k$.
The average $\frac{\sum_{k=1}^n x_k}{n}$ of numbers in $[0.5, 2]$ lies in $[0.5, 2]$ as well. 

As the function $f: x \in \mathbb{R^+} \mapsto x^n$ is convex, one has, for any $0 \leqslant a \leqslant x \leqslant b$, that $f(x)$ is below the line from $f(a)$ to $f(b)$:
$$f(x) \leqslant \frac{f(b) - f(a)}{b-a} (x-a) + f(a)$$

For our case of study, $a=0.5$ and $b=2$, this yields:
$$\forall x \in [0.5, 2], \;\; x^n \leqslant \frac{2^n - \frac{1}{2^n}}{2-0.5} (x-\frac{1}{2}) + \frac{1}{2^n}$$
that is,
$$\forall x \in [0.5, 2], \;\; x^n \leqslant \frac{2}{3} (2^n - 2^{-n}) (x-\frac{1}{2}) + 2^{-n}$$

$$\prod_{k=1}^n x_k \;\;\leqslant\;\; \left( \frac{\sum_{k=1}^n x_k}{n} \right)^n
 \;\;\leqslant\;\; 
 \frac{2}{3} (2^n - 2^{-n}) \left(\frac{\sum_{k=1}^n x_k}{n} -\frac{1}{2}\right) + 2^{-n}$$

 \subsection{Lower bound}

Let us denote by $f$ the product:
$$f: (x_1, x_2, \dots, x_n) \mapsto \prod_k x_k$$
Then note that at the middle point $(x_1, x_2, \dots, x_n) = (1,1,\dots,1)$:
$$\forall k, \;\; \frac{\partial f(x_1, x_2, \dots, x_n)}{\partial x_k} = \prod_{j\neq k}^n x_j = 1$$
and that consequently around the middle point, the first order approximation of the function is:
$$f(x_1, x_2, \dots, x_n) = f(1 + (x_1-1), 1+ (x_2-1), \dots, 1 + (x_n-1))$$ 
$$ = f(1,1,\dots,1) + \sum_k \frac{\partial f}{\partial x_k} (x_k-1) + O\left( (x_k-1)^2 \right)$$
$$= 1 + \sum_k (x_k-1) + O\left( (x_k-1)^2 \right)$$
$$= 1-n + \sum_k x_k + O\left( (x_k-1)^2 \right)$$
so that the linear function $(x_1, x_2, \dots, x_n) \mapsto 1-n + \sum_k x_k$ looks like a promising approximation of the function.
Unfortunately, as said earlier, the multiplication $f$ is not convex nor concave, so some parts of the graph of the function are above it and some other ones below.
Let us just remember that the hyperplane direction $\sum_k x_k$ sounds reasonable.

The tautology:
$$\prod_k x_k - \sum_k x_k \;\;\geqslant\;\; \inf_{y_1,y_2,\dots,y_n \in [0.5,2]} \left( \prod_k y_k - \sum_k y_k \right)$$
leads to:
$$\forall x_1, x_2, \dots, x_k \in [0.5, 2],$$  $$\prod_k x_k \geqslant \sum_k x_k + \inf_{y_1,y_2,\dots,y_n \in [0.5,2]}\left( \prod_k y_k - \sum_k y_k \right)$$

which leads us to a lower bound of the form:
$$\prod_k x_k \geqslant \sum_k x_k + C$$
for some constant $C$ that may depend only on $n$ and the interval chosen $[0.5 ,2]$.

Let us study the function:
$$g: (x_1, x_2, \dots, x_n) \in [0.5, 2]^n \mapsto \prod_k x_k - \sum_k x_k$$
We want to find its minimum over $[0.5, 2]^n$.
For each variable $x_k$: **if the minimum is reached in the interior of $[0.5, 2]$** (i.e. not at $x_k = 0.5$ or $2$), then necessarily at that point the derivative is 0:
$$\frac{\partial g}{\partial x_k} = \prod_{j\neq k} x_j - 1 = 0$$
i.e. 
$$\prod_{j\neq k} x_j = 1$$
and consequently $\prod_{j} x_j = x_k$.

Otherwise, \emph{if the minimum is reached on the boundaries of $[0.5, 2]$, then
either} $x_k = 0.5$ \emph{or} $x_k = 2$. 

For each $k$ we consequently have:
\begin{itemize}
    \item  either $\prod_{j\neq k} x_j = 1$
    \item or $x_k = 0.5$
    \item or $x_k = 2$
\end{itemize}

Note that if a variable $x_k$ satisfies the first property then:
$$g(x_1, x_2, \dots, x_n) = \prod_j x_j - \sum_j x_j = x_k - \sum_j x_j = - \sum_{j\neq k} x_j$$
which does not depend on $x_k$.
Thus in that case one can choose to change $x_k$ for $0.5$ or $2$ and this will not change the value of $g$.
Thus one can assume that all $x_k$ are $0.5$ or $2$, that is, the minimum is reached on a corner of the domain $[0.5, 2]^n$.

Let us assume that $K$ variables $x_k$ are $0.5$ and the $n-K$ remaining ones are $2$. Then:
$$g(x_1, x_2, \dots, x_n) = 2^{n-K} 0.5^K - ( (n-K)\, 2 + K\, 0.5))$$
i.e.
$$g(x_1, x_2, \dots, x_n) = 2^{n-2K} + \frac{3}{2} K - 2n $$
What is the value of $K\in[[0,N]]$ that minimizes this?

Let is study the function $h: x \in [0,N] \mapsto 2^{n-2x} + \frac{3}{2}x$. If it reaches a minimum strictly inside $[0,N]$ then at that point its derivative is 0:
$$-2\times 2^{n-2x} + \frac{3}{2} = 0$$
that is
$$\frac{2^{n+2}}{3} = 2^{2x}$$
$$x = \frac{1}{2}(n+2 - \frac{\log 3}{\log 2})$$
that is
$$x \simeq \frac{n}{2} + 0.2$$
This point is a minimum indeed (and not a maximum) as the second derivative of $h$ is positive.
Therefore the $K$ that we are searching for is the closest lower or upper integer to $\frac{n}{2} + 0.2$.

If $n$ is even: these are $\frac{n}{2}$ and $\frac{n}{2} + 1$. 

If $n$ is odd: these are $\frac{n-1}{2}$ and $\frac{n+1}{2}$.

By computing the associated values of $h$, one finds that the minimum in the even case is reached for $K = \frac{n}{2}$ and is $1+\frac{3}{4}n$,
while in the odd case, the same value is obtained for both possible values of $K$ and is $2 + \frac{3}{4}(n-1)$.

As $g = h - 2n$ at corners, this leads to:
- $\inf g = 1 - \frac{5}{4}n$ if $n$ is even
- $\inf g = \frac{5}{4} - \frac{5}{4}n$ if $n$ is odd

$$\forall x_1, x_2, \dots, x_k \in [0.5, 2],$$  $$\prod_k x_k \geqslant \sum_k x_k + 1 - \frac{5}{4}n  + \frac{1}{4}\delta_{n \text{ is odd}}$$

with $\delta_{n \text{ is odd}} = 1$ if $n$ is odd and $0$ otherwise.
The bound is tight and reached on many corners (all the ones with half lowest and half highest coordinates) as well as on the edges linking these corners if $n$ is odd (free variable that can take any value).

Final result:

$$\forall x_1, x_2, \dots, x_k \in [0.5, 2],$$  $$\sum_k x_k + 1 - \frac{5}{4}n  + \frac{1}{4}\delta_{n \text{ is odd}} \;\;\leqslant\;\; \prod_k x_k$$
$$\prod_k x_k \;\;\leqslant \;\;  \frac{2}{3} (2^n - 2^{-n}) \left(\frac{\sum_{k=1}^n x_k}{n} -\frac{1}{2}\right) + 2^{-n}$$

\end{document}